 \documentclass[final,3p,times,twocolumn]{elsarticle}

\usepackage{amssymb}

\usepackage{url}
 \usepackage{amsmath} 
\usepackage{hyperref}

\journal{Neural Networks}

\begin{document}

\begin{frontmatter}



\title{High Speed Human Action Recognition using a Photonic Reservoir Computer\\}

 \author[label1]{Enrico Picco\corref{cor1}}
\ead{enrico.picco@ulb.be}
\cortext[cor1]{Corresponding author}

 \affiliation[label1]{organization={Laboratoire d’Information Quantique, CP 224, Université Libre de Bruxelles (ULB), B-1050, Bruxelles, Belgium}
             }
 \author[label2]{Piotr Antonik}
 \affiliation[label2]{organization={MICS EA-4037 Laboratory, CentraleSupélec, F-91192 Gif-sur-Yvette, France\\}
             }
 \author[label1]{Serge Massar}

\begin{abstract}
The recognition of human actions in videos is one of the most active research fields in computer vision. The canonical approach consists in a more or less complex preprocessing stages of the raw video data, followed by a relatively simple classification algorithm. Here we address recognition of human actions using the reservoir computing algorithm, which allows us to focus on the classifier stage. We introduce a new training method for the reservoir computer, based on “Timesteps Of Interest”, which combines in a simple way short and long time scales. We study the performance of this algorithm using both numerical simulations and a photonic implementation based on a single non-linear node and a delay line on the well known KTH dataset. We solve the task with high accuracy and speed, to the point of allowing for processing multiple video streams in real time. The present work is thus an important step towards developing efficient dedicated hardware for video processing.
\end{abstract}




\end{frontmatter}


\section{Introduction}

Human Action Recognition (HAR) is nowadays considered a milestone in the field of Computer Vision (CV). During the past decades it has become a prolific research area due both to the importance of 
video processing  in everyday life and to the tremendous advances 
in artificial intelligence.
Initially considered  for surveillance and human-machine interaction \cite{Moeslund}, HAR also finds use in areas such as autonomous driving, medical rehabilitation, content-based video indexing on social platforms, as well as education \cite{Singh, sun22}.  
To target these applications, it is necessary to offer reliable  autonomous scene-analysis using embedded hardware to process data in real time and with high energy efficiency. However HAR remains a challenging task due to the nature of the actions which must be recognized, as well as the varying contexts in which they take place involving for instance variations in lighting and illumination, camera scale and angle variations, frame resolution, etc.

The usual approach of Computer Vision systems for HAR recognition is 
to separate the optimization into two major steps \cite{Ji2012}. The first step consists in providing a more concise and easily parsed representation of the video. To this end preprocessing techniques and algorithms are used to extract a set of relevant characteristics (also called features) such as the various homogeneous regions present in the image, texture, movement, color, etc. The preprocessing step has been the main topic of research in this area.
The second step is to classify 
the preprocessed data with more complex  algorithms which allow the analysis and identification of activities in video sequences \cite{Wiley}. In this respect deep learning is by now well established as a powerful classification tool for human action recognition \cite{Wu}. However it comes with several drawbacks such as time-consuming and power-inefficient training procedures, and the need for dedicated high-end hardware.
For a review of  different preprocessing and classification techniques, see  \cite{Singh, Poppe, AbuBakar}. 

One of the most widely used datasets for HAR is the KTH dataset, first introduced in \cite{Schuldt, Laptev} and then used in hundreds of publications \cite{Chaquet} such as \cite{Sharif, Khan, Rahman, Rathor, Shu, Jahagirdar, Jhuang, Ramya, Grushin, Ji, Xie, Liu, Begampure, AntonikHuman}; the reader can refer to \cite{Singh} for a detailed comparison between some relevant works on this dataset.

In this paper, we adopt a different approach, focusing our attention on the classifier stage.
To this end we use Reservoir Computing (RC) for the HAR task, both in software and in an opto-electronic implementation. 
Reservoir Computing is a set of Machine Learning algorithms for designing and training artificial neural networks \cite{Jaeger, Maass,Lukosevicius} that are highly efficient for processing time series.  
Reservoir Computers are significantly easier to train than other Neural Networks. Indeed, they consist of an untrained random recurrent neural network which is driven by the time series to analyse, and a trained readout layer. 
 Reservoir computing has been applied to a variety of problems, including for instance speech recognition \cite{verstraeten2006reservoir} and time series prediction\cite{Jaeger,pathak2018model}.

Reservoir computers (RCs) have been  implemented physically using e.g.  spintronics \cite{Akashi, Torrejon}, laser-based systems, \cite{Akrout, Butschek, Vinckier, BrunnerParallel, Duport}, microresonators \cite{Bazzanella, Borghi}, free-space optics \cite{Bueno, Dong}, VCSELs \cite{Skalli, Vatin}, memristors \cite{Tong, Tanaka}, integrated photonics \cite{Vandoorne, Takano}, and optoelectronic systems \cite{Martinenghi, Paquot}.
These implementations have been successfully tested on benchmark tasks including hand-written digits \cite{AntonikLarge}, speech \cite{Triefenbach,LargerMillion} and video \cite{AntonikHuman, Lu}  recognition, prediction of future evolution of financial \cite{2006} and chaotic \cite{AntonikBrain} time series, with performances comparable to digital Machine Learning  algorithms.
Photonic RC in particular is a promising path towards the development of real-time and energy-efficient information processing systems. 

As a video sequence is  a time domain signal, classifying video signals seems particularly well adapted to the Reservoir Computing approach. Furthermore, classifying video sequences  is rather complex and will therefore test the performance of RC on a more challenging task than often used in the literature.

RC implementations have been used previously to target the  HAR task. Zheng et al. report using microfabricated MEMS resonators for HAR \cite{Zheng_2022}. However, they reported problems in reservoir size scalability and limited the size of their implementation to N=200 nodes. Lu et al. use micro-Doppler radar signal processing for HAR \cite{Lu}. However, they achieved classification accuracy going only up to 85\% which is behind the state-of-the-art  for HAR. Antonik et al. used a large-scale photonic computer using free-space optics. Their implementation can scale the size of the reservoir from several tens to hundreds of thousands of nodes \cite{AntonikHuman}. However, their system was rather slow, and free-space optical experiments can suffer from environmental influences such as noise and temperature variations.  

Here we apply the RC architecture based on a delay loop and a single nonlinear node to HAR. This architecture has been shown to provide excellent performance while being simple to implement, both in software \cite{Rodan} and in hardware \cite{Paquot, LargerMillion, Appeltant,LargerTuring}. 
To benchmark our results, we use the aforementioned KTH dataset.

On the conceptual side, our main advance is the introduction of a novel architecture for the output layer, inspired by Ref.~\cite{AntonikLarge,Schaetti} where a similar approach was used for static image recognition, which we call using {\em Timesteps Of Interest} (TOI). This allows to increase the dimensionality of the output layer without modifying the size of the reservoir, while simultaneously introducing a new time scale which depends on which timeframes are used. 

We report results on an optoelectronic RC and on simulations thereof. The  optoelectronic RC is based on off-the-shelf optical components and a Field Programmable Gate Array (FPGA) board, following our previous works (see e.g.  \cite{AntonikBrain}). 
Our study shows that, despite its simplicity, a small-scale delay RC is  capable of solving the complex HAR task, achieving performance comparable to the state-of-the-art digital approaches. We experimentally achieved a classification accuracy value of 90.83\% for a reservoir size of N = 600. Because of their simplicity, both the numerical and experimental implementations are considerably faster than previous systems, reaching more than $100$ frames per second (fps). Our work therefore provides a very promising approach for real time, low energy consumption, video processing.

Our paper is structured as follows. We first introduce the KTH database and the dataprocessing pipelines usually used for HAR. We then introduce Reservoir Computing and the concept of Timesteps Of Interest, followed by a description of our experimental system. In the results section we present classification acccuracy as a function of the size of the reservoir, the number of TOI, the input data variability kept after Principle Componennt Analysis (PCA), and we compare our results with those obtained in previously in the literature. 

\section{Data Processing and the Experiment}

\subsection{Human Action Recognition}

\subsubsection{KTH Human Actions Dataset}
\begin{figure}[h]
\includegraphics[width=8cm]{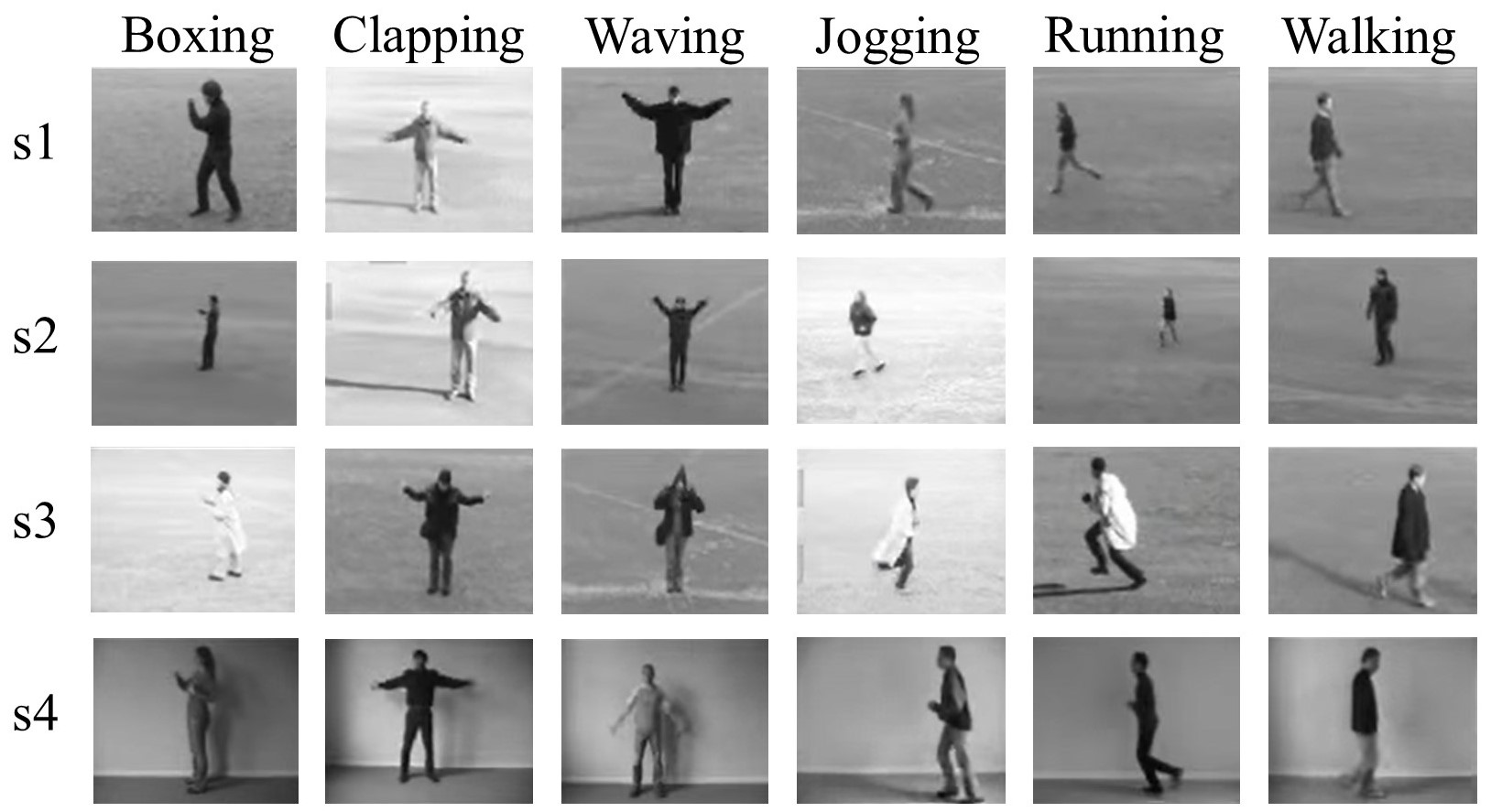}
\caption{Samples of the KTH Human Actions dataset showing six human actions output classes and four scenarios (s1: outdoors, s2: outdoors with scale variation, s3: outdoors with different clothes, s4: indoors) \cite{Schuldt}}
\label{fig:KTH}
\end{figure}

The KTH Human Actions dataset is provided by the Royal Institute of Technology in Stockholm \cite{Schuldt, Laptev}. It constitutes a well-known benchmark in the Machine Learning community to compare the performance of different learning algorithms for recognition of human actions \cite{Poppe, AbuBakar}. The dataset consists of video sequences divided into six different classes of human actions, namely walking, jogging, running, boxing, hand waving, and hand clapping. All the sequences are stored using AVI file format and are available online (\url{https://www.csc.kth.se/cvap/actions/}). The actions are performed by 25 subjects recorded in four different scenarios, outdoors s1, outdoors with scale variation s2, outdoors with different clothes s3, and indoors s4. Each action is repeated 4 times by each subject. The video sequences are taken with a 25fps frame rate, down-sampled to the spatial resolution of 160 x 120 pixels, and have a time length of 4 seconds on average. Their length varies between 24 and 362 frames \cite{Schuldt}. Fig.~\ref{fig:KTH} shows a few samples of the data.

Due to some missing sequences, the original dataset contains a total of 2391 video sequences, instead of 25×6×4×4 = 2400 sequences for every combination of 25 subjects, 6 actions, 4 scenarios, and 4 repetitions. For ease of use, we increase the number of sequences to 2400 by filling in missing action sequences with a copy of the existing ones taken from the same subject and scenario. Consequently, the modified dataset used in this study contains 2400 video sequences. The sub-datasets corresponding to four different scenarios contain 600 videos each.

\subsubsection{Data Pre-Processing}\label{sec:pre-processing}
\begin{figure}[h]
\includegraphics[width=8cm]{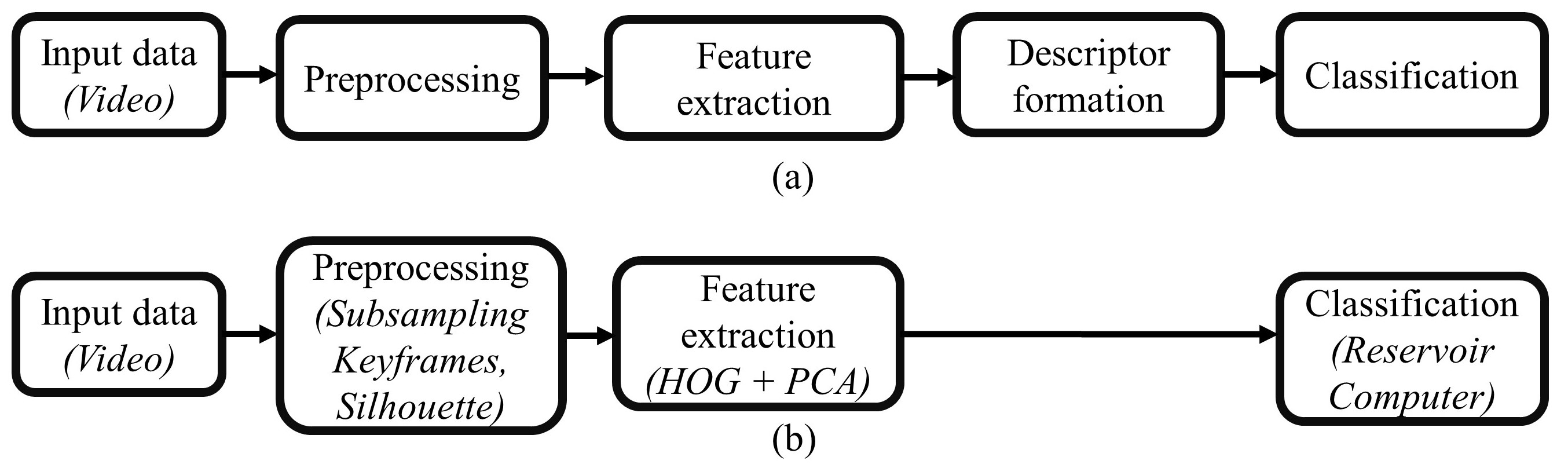}
\caption{Process flow in human action recognition: (a) canonical approach (adapted from \cite{AbuBakar}); and (b) our approach.}
\label{fig:preprocess}
\end{figure}

Usually, the deep-learning-based approach for human-action recognition consists of the following steps: (i) the preprocessing of the input data that might include, for instance, the removal of the background, (ii) feature extraction via a series of convolutional and pooling layers, (iii) formation of the feature descriptors to numerically encode interesting information, and (iv) the actual classification using various classification algorithms (cf. Fig. \ref{fig:preprocess}(a)) \cite{AbuBakar}. In this chain of actions, the formation of feature descriptors is realised using a feature-description algorithm and can be a time- and calculation-consuming task. As the reservoir computer is per se designed to process time-dependent signals, a separate descriptor-formation stage is not necessary. 
This simplifies the overall data processing workflow and contributes to the increase of the data processing speed (cf. Fig. \ref{fig:preprocess}(b)) .

In more detail the preprocessing and feature extraction we use consists of the following steps. First we subsample each video sequence to extract 10 keyframes to allow the classification to work on an equal number of frames for each video sequence (\ref{app_sub}). Then, we use an attention mechanism to generate binary silhouettes by discarding the background and horizontally centering the silhouettes in the frame \cite{LuSemantic} (\ref{app_sil}). This removes any unnecessary information (such as background clutter or shadows) and focuses the attention of the classifier on the pose of the subject disregarding its spatial position in the frame. The feature extraction step consists in the computation of the histograms of oriented gradients (HOG)  to automatically detect the silhouettes in the frames (\ref{app_feature}) \cite{Dalal}. Finally, to simplify the computation, we reduced the number of HOG features using the principal component analysis (PCA) (cf. Sec. \ref{sec:PCA}) \cite{Pearson, Hotelling, Smith}. The resulting features are then injected into RC for classification.

\subsection{Reservoir Computing}

\subsubsection{Basic Principles}\label{sec:basic}
In general form, a RC contains $N$ internal variables $x_i (n)$ that are often named “nodes” or “neurons” as derived from the biological origins of artificial neural networks. The nodes are concatenated in a state vector $x(n) = x_{i\in0...N-1}(n)$. The evolution of the neurons in discrete time  $n \in Z$ can be expressed as
\begin{equation}
x(n+1) = f(Wx(n) + W^{in}u(n))
\label{eq:RC}
\end{equation}
where $f$ is a nonlinear scalar function, $W$ is an $NxN$ matrix representing the interconnectivity between the nodes of the network, $u(n)$ is the $K$-dimensional input signal injected into the system, and $W^{in}$ is the $NxK$ internal weight matrix, also called mask. The entries of $W$ and $W^{in}$ are time-independent coefficients with values drawn from a random distribution with zero mean and variance  adjusted to determine the dynamics of the reservoir and obtain the best performance on the investigated task.

Here, we use a ring-topology reservoir with a nearest-neighbor coupling between the nodes.
Although quite simple, this architecture has shown equivalent performance to more complex reservoir implementations, both numerically 
\cite{Rodan} and experimentally \cite{BrunnerParallel, Paquot, LargerTuring, Appeltant, DuportFully}. The nonlinearity that we use here is sinusoidal (cf. \cite{Paquot, LargerTuring}) and can be easily implemented experimentally \cite{AntonikPattern} (cf. Sec. \ref{subsubsection:exp}). 
In terms of mathematical description, our topology changes Eq. \ref{eq:RC} to: 
\begin{equation}
\begin{aligned}
& x_0(n+1) = \sin(\alpha x_{N-1}(n-1) + \beta M_0u(n)) \\
& x_i(n+1) = \sin(\alpha x_{i-1}(n) + \beta M_iu(n)) & i = 1,.., N-1 
\end{aligned}
\label{eq:ring}
\end{equation}
The interconnectivity matrix W is now reduced to a single feedback parameter $\alpha$.  
The input term is calculated by multiplication of a NxK mask $M_i$ with a K-dimensional input signal $u.$ The input strength parameter $\beta$ tunes the reservoir dynamics .

\begin{figure}[h]
\includegraphics[width=8cm]{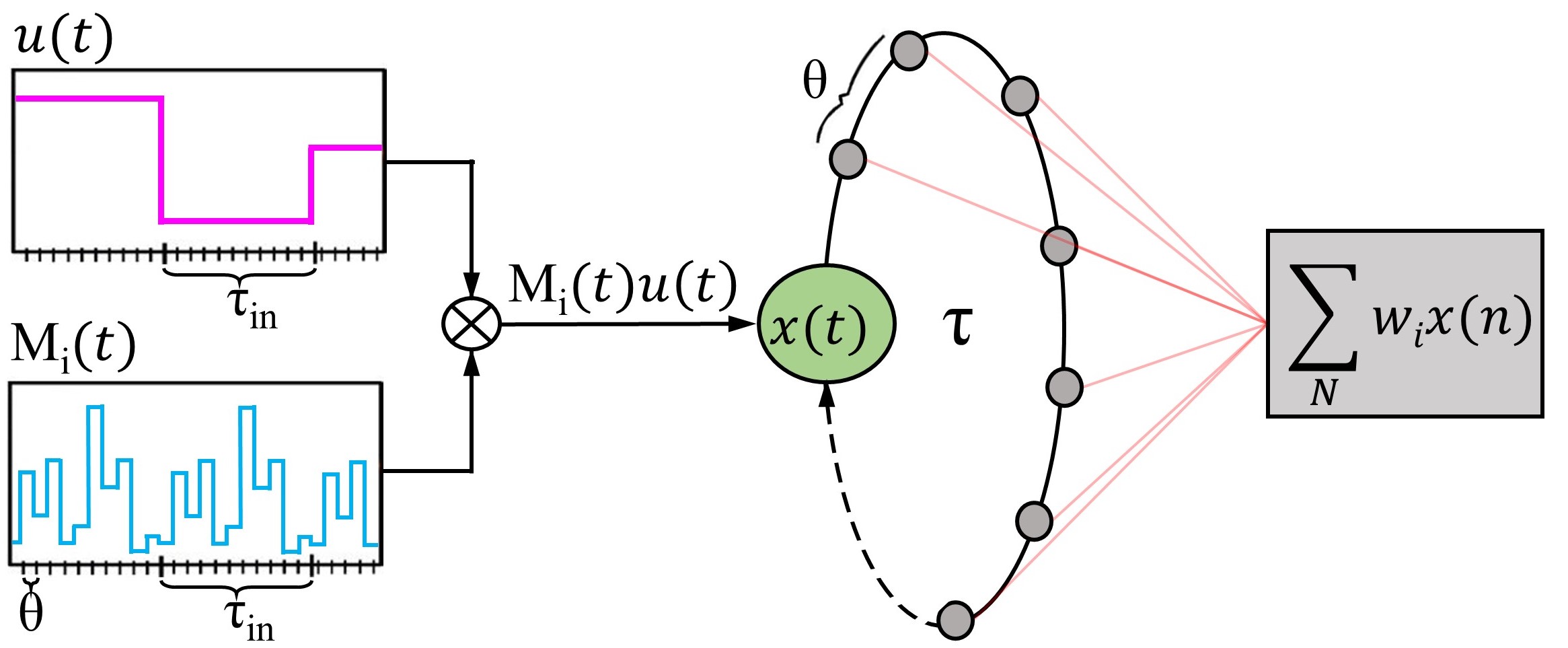}
\caption{Basic scheme of a time-multiplexed reservoir computer. The product of input $u(t)$ and mask $M_i(t)$ generates the masked input that drives the reservoir states $x(t)$, which are then used for training and testing. }
\label{fig:timedelay}
\end{figure}

In this work, the RC is implemented using an optical fiber loop with time delay $\tau$, illustrated schematically in Fig.~\ref{fig:timedelay}. 
 In this work we use desynchnization between the input and the reservoir, as introduced in \cite{Paquot}
 To this end the input signal $u(t)$ is sampled and held for a period of time $\tau_{in}$ which is different from the delay loop period $\tau$. Desynchronization is a technique for coupling the neurons in the reservoir and create interaction between them: without desynchronization, the system would not be a network of interconnected neurons, but only a set of independent variables.
In the present work the time duration of one neuron is taken to be $\theta=\tau / (N+1)= \tau_{in}/N$. It follows that $\theta$ is the time difference between $\tau$ and $\tau_{in}$.
After sample and holding, the input signal is multiplied by the mask $M_i(t)$, a periodic signal of period $\tau_{in}$, with each period containing $N$ random values of duration $\theta$ taken over the interval $[-1, +1]$. The masked input signal $M_i(t)u(t)$ is then injected into the reservoir, i.e. injected into the fiber spool. 

The output of the reservoir $y(n)$ is given by
\begin{equation}
y(n) = \sum_{i=0} ^{N} w_ix_i(n)
\label{eq:y}
\end{equation}
where $w_i$ are the weights of the linear readout layer. 

Contrary to the traditional RNNs where all interconnections between the neurons are trained, RC only trains its output layer, making the computation  faster and always convergent \cite{Lukosevicius, LargerMillion}. 
The dataset is split in a train set and a test set. In both train and test phase, the system is driven by the input $u(t)$ to excite the reservoir states $x(t)$. Furhermore, in this work we reset the reservoir states prior to each input video sequence, to enhance the performance: more considerations and quantitative analysis on the impact of reservoir state reset can be found in \ref{app_reset}. The reservoir states related to the training set are then stored in the train state matrix $X(t)=(x(t_1), ... , x(t_{K_{tr}})),$ $X \in R^{NxK_{tr}}$ with $K_{tr}$ is the size of the train dataset. The most often used training method to obtain the weights $w_i$ is regularized linear (ridge) regression \cite{Tikhonov}:
\begin{equation}
w = (X^TX+\lambda I)^{-1}X^T \tilde y
\label{eq:ridge}
\end{equation}
where $\lambda$ is the regularization parameter and $\tilde y$ is the desired output. 

For classification problems, such as speech recognition or HAR, the usual approach  \cite{Verstraeten} is to train as many outputs output $y_j(n)= w_j x(n)$, $j=1,...,C$, as their are output classes $C$, with $w_j$  the output weights for class $j$. 
If the current input $u(n)$ belongs to class $c\in C$, then $y_c(n)$ is trained to take a high value, say $1$, and the other outputs $y_j(n)$ ($j\neq c$) are trained to take a low value, say $0$. 
Thus for the HAR task, as there are 6 classes, one would train 6 sets of output weights to take either $0$ or $1$ values.
During the test phase, one computes which output, averaged over the duration of the input, has the highest value, and the input is assigned to the corresponding class, through a winner-takes-all approach.

\subsubsection{Training using Timesteps Of Interest}

In the present work we use a novel training and classification procedure which is inspired by Ref.~\cite{AntonikLarge,Schaetti} where a similar approach was used for static image recognition. However, the method has never been applied to a temporal task like video stream processing. Here the concatenated states are related to different frames of the video, rather than different sections of a static image.
 
To introduce the method,  recall that the input is subsampled to contain $10$ keyframes. That is both the input $u(n)$ and  all of the reservoir states $x_i(n)$ have a duration of length $10$: $n=1,...,10$.
From these $10$ time steps, we select a subset which we call the {\it Timesteps Of Interest} (TOI). For instance these could consist of time frames $2,8,10$. We then concatenate the reservoir states at the time frames of interest. In the above example this would yield a vector $X$ of size $3N$ equal to $X=(x(2), x(8), x(10))$.

\begin{figure*}
\includegraphics[width=16cm]{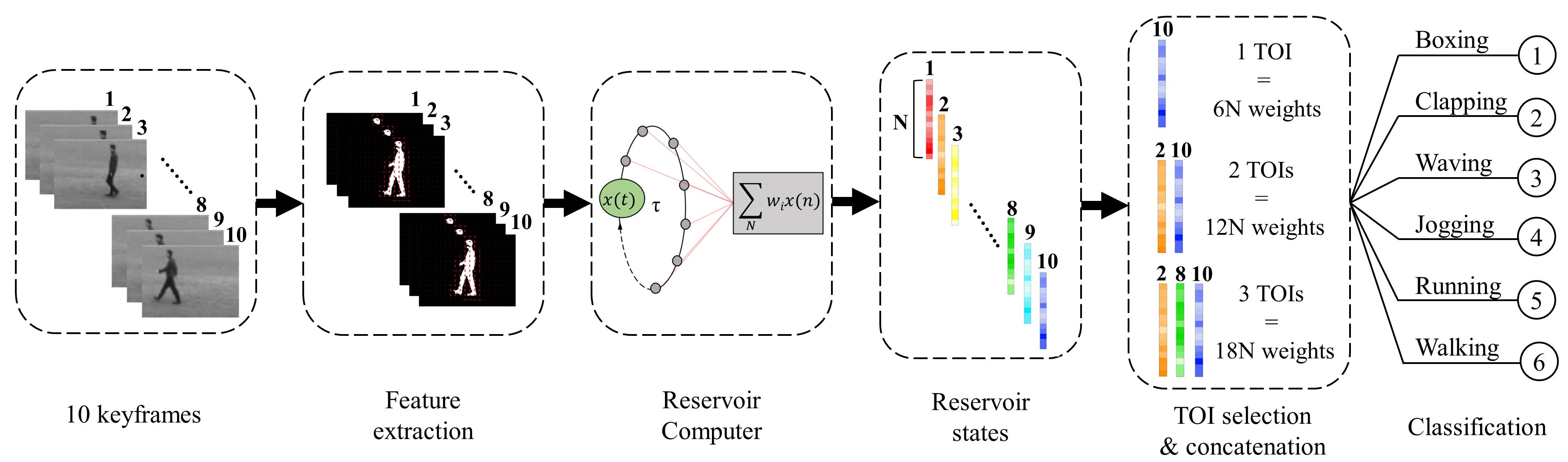}
\caption{Illustration of the use of reservoir computer for human action recognition. First, 10 keyframes are selected from the subsampled video sequence, then histograms of oriented gradients (HOG) are extracted from binary (black-and-white) silhouettes to generate the input features (see Apps. A-C). After injection in the RC, each keyframe induces one state vector of the network. The state vectors at the Timesteps Of Interest  are concatenated into one vector. In the figure we illustrate this for the cases where a single TOI is used (corresponding to time $n=10$), two TOI are used (corresponding to times $n=2$ and $n=10$, and three TOI are used (corresponding to times $n=2, 8, 10$. Six outputs (one for each output class) are trained, corresponding to $6N$, $12N$, or $18N$ output weights depending on how many TOI are used. Finally the input is assigned to the output class whose corresponding value is largest.
}
\label{fig:flowandTOI}
\end{figure*}

We then train as many outputs as their are classes (in the present case 6 outputs corresponding to the 6 classes of the HAR task)
\begin{equation}
    y_j = w_j X
\end{equation}
where the $w_j$ are vectors of size $3N$ (since $3$ is the number of time frames of interest in the above example). The $w_j$ are trained using ridge regression, see Eq.  (\ref{eq:ridge}), so that if the current input $u(n)$ belongs to class $c\in C$, then $y_c$ take a high value, say $1$, and the other outputs $y_j$ ($j\neq c$)  take a low value, say $0$. During the test phase, one computes all the  outputs $y_j$ and assigns the input to the class $c$ whose output has the highest value, through a winner-takes-all approach. This classification procedure is illustrated in Fig.~\ref{fig:flowandTOI}.

A first advantage of the present method is that it increases the number of output weights $w_j$, i.e. the number of trained parameters, without increasing the size of the reservoir. A second advantage is that the reservoir output 
now takes into account both the short term memory present in the reservoir itself, and a long term memory which is implemented through the concatenation of the key frames of interest at selected times. 
Generally speaking, one of the trade-offs in tuning a reservoir's hyperparameters involves memory and nonlinearity \cite{Verstraeten2010}: linear memory capacity decreases for increasingly nonlinear activation functions. Here, by combining short-term and long-term memories with the TOIs, we artificially achieve a highly nonlinear system with long memory, thus avoiding the need to compromise.

The disadvantage of the present method is that one needs to choose the TOI.  Indeed, since increasing the number of TOI 
increases the complexity of the output layer, one will try to choose the minimum number of TOI compatible with good performance. 
The benefits of this technique and the optimal number of training TOIs are discussed further in \ref{sec:TOIs}. Overall, using the TOI yields an increase in performance, as we demonstrate below.

\subsubsection{Hyperparameters and Bayesian Optimization}\label{sec:hp}
The RC dynamics depend on task-dependent coefficients called hyperparameters. The choice of these coefficients is a critical step in the design of a RC. In order to maximize the RC performance, the hyperparameters need to be tuned for every benchmark task. Our RC has three hyperparameters that need to be optimized:
\begin{itemize}
  \item the feedback parameter $\alpha$ represents the strength of the recurrence of the network.
  \item the input strength $\beta$ is the scale factor of the input  in Eq.~\ref{eq:ring}. It defines the amplitude of the input signal and, thus, the degree of nonlinearity of the system response. The value of $\beta$ by itself is not meaningful, as it depends on the scale used for the input $u(n)$. In the table below we give the standard deviation of the input $\sigma_{\beta M_i u(n)}=\sqrt{{\text VAR} (\beta M_i u(n))}$.
 \item the ridge parameter $\lambda$ is the coefficient used in Eq.~\ref{eq:ridge}, it adds regularization to the training of the linear regression model. The value of $\lambda$ by itself is not meaningful, as it depends on the scale used for the neurons $x_i(n)$. In the table below we give the rescaled value $\lambda /{\text VAR} (x(n))$.
\end{itemize}

In the RC community, the hyperparameters are usually determined via a grid search. The grid search tests every possible combination of hyperparameters over an interval with a pre-defined precision. This method is simple, but the processing time quickly explodes in the case of slow experimental setups and large number of hyperparameter combinations.
To counterbalance the problem of exploding processing time, we use Bayesian optimization \cite{Mockus, Brochu} to find optimal parameters. Its effectiveness in the case of RCs has been shown in Ref.~\cite{AntonikBayesian}.
This method creates a surrogate model of the classification accuracy as a function of the hyperparameters using Gaussian Process regression \cite{Rasmussen}. The hyperparameter space is then efficiently scanned by an acquisition function by making a trade-off between the exploration and exploitation of the regions with more potential for improvement (cf. \ref{app_bayes}). To compare, a grid search would need a few thousand iterations to find the optimum hyperparameters in our setup whereas the Bayesian optimization takes less than a hundred iterations to converge. In practical terms, finding optimal hyperparameters for our experimental system reduces from 3 weeks (grid search) to roughly half a day (Bayesian optimization).

In our simulations and experiments, given the limited size of the dataset, the hyperparameters were optimised on the whole dataset instead of using a dedicated validation set. Table \ref{table:hp} gives the optimum hyperparameters found for our simulations and for the experiment, for two different reservoir sizes ($N=200$ and $N=600$). In experiments, the attenuation in the fiber loop corresponds to the hyperparameter $\alpha$ (cf. Sec. \ref{subsubsection:exp}), but it is not immediately related to the value of  $\alpha$. For this reason, the optimal experimental values of $\alpha$ are not included in the table. The values given in Table  \ref{table:hp}  were used throughout this work when other parameters (such as the number of TOI) were varied. Note that the optimal rescaled values of beta and lambda are not the same in the simulations and experiment. In the case of the ridge parameter lambda, this is presumably because the experiment is affected by noise.

\begin{table}[h]
\centering
\resizebox{8cm}{!}{
\begin{tabular}{cc|ccccc}
  & N & $\alpha$ & $\beta$ & $\sigma_{\beta M u}$ & $\lambda$ & $\lambda /{\text VAR} (x)$ \\ \hline
Simulation & 200 & 1.90 & 0.0004 & 0.0016 & 0.00013 & 0.0029  \\ 
 & 600 & 1.50 & 0.0032 & 0.014 & 0.22 & 4.07 \\ \hline
Experiment & 200 & / & / & 0.078 & / & 68.29  \\ 
 & 600 & / & / & 0.065 & / & 16.57 \\ 
\end{tabular}
}
\caption{ Optimal hyperparameters for simulation and experiment for different number of neurons. These values are used throughout this work.}
\label{table:hp}
\end{table}

The rescaled values $\sigma_{\beta M u}$ and $\lambda /{\text VAR} (x)$ do not depend on the scales used for the input and internal variables. For the experimental implementation, we do not give the values of $\alpha$, $\beta$, $\lambda$ as these depend on the scales used.

\subsubsection{Experimental  Reservoir Computer}\label{subsubsection:exp} 
\begin{figure}[h]
\includegraphics[width=8cm]{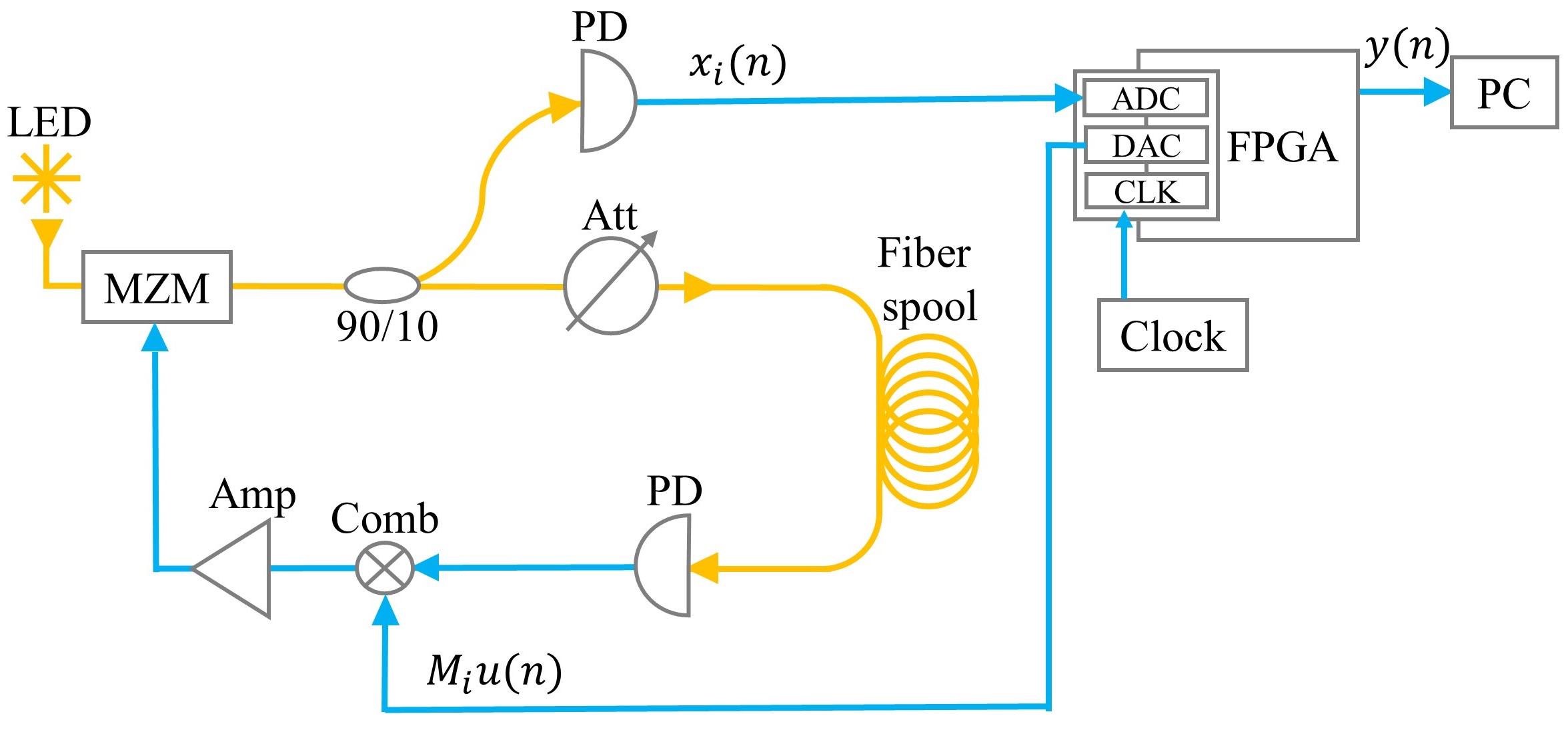}
\caption{Schematic of the optoelectronic experimental setup. The photonic part of the reservoir is shown in yellow and  is composed of an incoherent light source (LED), a Mach-Zehnder Modulator (MZM), an optical attenuator (Att), a fiber spool, and two photodiodes (PD). The electronic part is shown in blue and consists of an FPGA board connected to a PC, a clock generator (Clock/CLK) to drive the analog-to-digital converter (ADC) and digital-to-analog converter (DAC), a resistive combiner (Comb), and an amplifier (Amp). }
\label{fig:schematic}
\end{figure}

Our experimental setup is shown in  Fig.~\ref{fig:schematic}. It consists of two main parts: an optoelectronic reservoir (cf. Fig.~\ref{fig:timedelay}) and an FPGA board (cf. \cite{Paquot, AntonikBrain, LargerTuring}). The optoelectronic part starts with a superluminescent diode (Thorlabs SLD1550P-A40). The output of the diode passes through a Mach-Zehnder intensity Modulator (MZM) (EOSPACE AX-2X2-0MSS-12) to implement the sinusoidal nonlinearity of Eq.~\ref{eq:ring}. A 10\% fraction of the MZM output is sent to a photodiode (TTI TIA-525I) for the neurons' readout. The remaining 90\% is sent to an optical attenuator (JDS HA9) to tune the feedback parameter $\alpha$ in Eq.~\ref{eq:ring}. 

The ring-topology of the reservoir is implemented using time-multiplexing: the delay that the light needs to travel through the fiber spool is divided by the desired number of neurons. This is done by selecting the  clock frequency (CLK) of the electronic part of the system so that every neuron is encoded in light intensity for a duration $\theta=\tau / (N+1)$, as discussed in section \ref{sec:basic}.

The neurons are  collected from the spool with a second photodiode, electrically combined with the masked input $M_iu(n)$, and sent back to the modulator after an amplification stage of $+27$ dB (ZHL-32A+ coaxial amplifier) to span the entire V\textsubscript{$\pi$} interval of the MZM. 
A power supply (Hameg HMP4040) provides the bias voltages for the amplifier and the MZM.

The electronic part of the setup is based on a FPGA board. Its reconfigurability allows us to flexibly try out different experimental configurations.
In our setup, the FPGA interfaces with the reservoir using an analog-to-digital converter (ADC) and a digital-to-analog converter (DAC), and also communicates with a PC. The ADC collects the states from the readout photodiode so that they can be used for training and testing, whereas the DAC sends the masked inputs $M_iu(n)$ to the reservoir. The FPGA-PC link is realized with a custom-designed PCIe interface that provides a higher bandwidth than off-the-shelf Ethernet-based solutions. A more detailed description of the FPGA design can be found in \ref{app_FPGA}.

The setup is tested with two different reservoir sizes, $N=200$ and $N=600$. To do so, we used two different fiber spools of respectively 1.6 km ($\tau = 7.94 \mu s$) and 10 km ($\tau = 49.27 \mu s$).
For the HAR task, each input video frame is a K-dimensional vector after preprocessing. Therefore, the mask $M_i$ is a matrix of size NxK.
In this way, the input-mask matrix product reshapes an 1xK input into a 1xN vector, ready to be fed to the N-sized reservoir.

\section{Results and Discussion}
Here, we present the numerical and experimental results of our system on the HAR task. We focus on how the reservoir size, number of TOIs, and dimensionality reduction after PCA influence the classification accuracy. In Sec.~\ref{sec:comparison}, our system is compared with other published schemes. 
In \ref{app_RC_role} we show that using a reservoir significantly improves performance compared to simply carrying out linear regression on the preprocessed inputs.

\subsection{Performance vs. Reservoir Size N}\label{sec:reservoirsize}
The reservoir size N is generally the most important parameter to take into account in the design of a RC.  Usually, a larger network has more computational capability and memory which results in better performance.

First, we investigate the impact of N (reservoir size / number of nodes) on the classification accuracy both, numerically and experimentally, using scenario s1 that comprises boxing, clapping, waving, jogging, running, and walking outdoors. (We focus only on one scenario to reduce the overall computational time).

\begin{figure}[h]
\includegraphics[width=8cm]{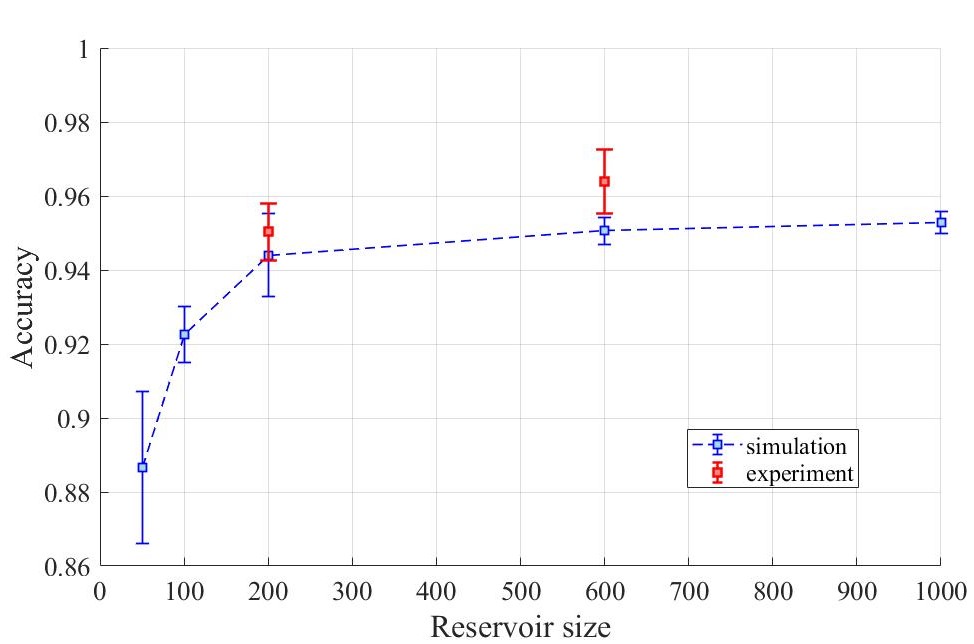}
\centering
\caption{Impact of the reservoir size N on its performance studied numerically (blue) and experimentally (red) for scenario s1 (boxing, clapping, waving, jogging, running, and walking performed outdoors), using 5 Timesteps of Interest. Error bars are standard deviation as described in the main text.
}
\label{fig:acc_vs_N}
\end{figure}

The network has been simulated for N=50, 100, 200, 600, and 1000. As seen in Fig.~\ref{fig:acc_vs_N}, the performance sharply increases with the network size of up to 200 nodes (blue squares). Then, it continues improving. Experimentally, we use an RC with N=200 and N=600 (cf. Sec.~\ref{subsubsection:exp}). The experimental results are depicted (red squares) in Fig.~\ref{fig:acc_vs_N}, and are in a good agreement with the numerical ones. 
 
For this figure, we ran the RC with 5 different input masks, and  used K-fold cross validation (with K=4) to make up for the limited size of the dataset \cite{Verstraeten}. The obtained results with their standard deviation are shown in Fig. ~\ref{fig:acc_vs_N}.

Next we performed experimental studies for all scenarios (s1-s4) as well as the whole dataset for $N=200$ and $N=600$. Tab.~\ref{table:accvsN} summarizes the results. As expected, the system performs better for a higher number of N for separate scenarios as well as for the whole dataset. The best classification accuracy we achieved experimentally on the complete dataset is 90.83\% for N=600.  
Tab.~\ref{table:accvsN} also shows that the video sequences of Scenario 2 are more difficult to classify because they are shot with different zoom and scale variations. This trend will be evident in the next sections.

\begin{table}[h]
\centering
\begin{tabular}{l|c c}
 \textbf{Dataset} & \textbf{N = 200} & \textbf{N = 600}\\
 \hline
 Scenario 1 & 95,33\% & 96,67\% \\
 Scenario 2 & 82,67\% & 87,33\% \\
 Scenario 3 & 94,00\% & 95,33\% \\
 Scenario 4 & 91,33\% & 92,00\% \\
\textbf{Full KTH dataset} & \textbf{86,83\%} & \textbf{90.83\%}\\
\end{tabular}
\caption{Experimental results on individual scenarios (s1, s2, s3, s4) and on the full database for two different reservoir sizes, N=200 and N= 600, using 5 Timesteps of Interest.}
\label{table:accvsN}
\end{table}

\subsection{Reservoir Computer Performance vs. Timesteps of Interest}\label{sec:TOIs}

\begin{figure}[h]
\includegraphics[width=8cm]{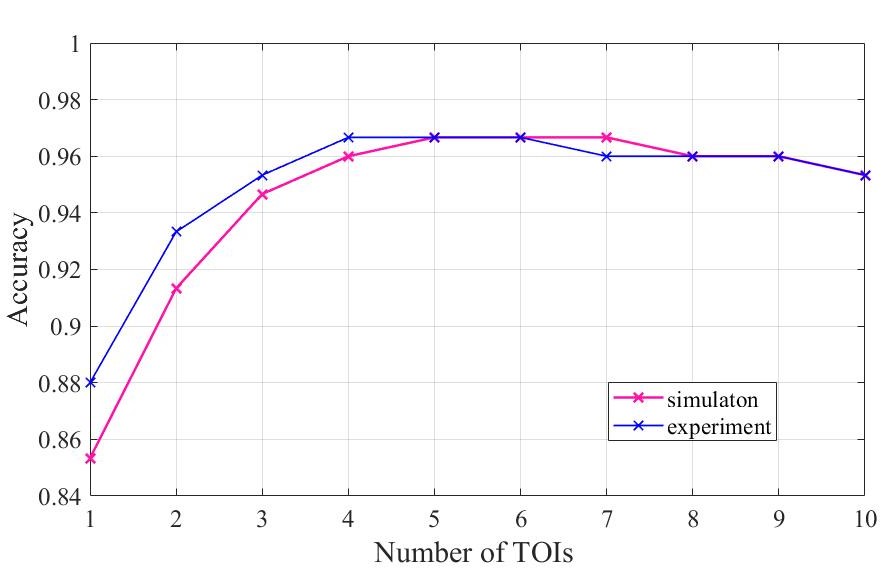}
\centering
\caption{Impact of number of TOIs on performances, for N=600, on scenario 1. Numerical results are in magenta and experimental results in blue. The presented results are for the optimal choice of TOIs.}
\label{fig:acc_vs_TOI_s1_0.75_N_200}
\end{figure}

The number of TOIs is the second major parameter to impact the classification performance. As we feed the reservoir with 10 keyframes for each video sequence, 10 is also the maximum number of TOIs that we can use. To find the optimal number of TOIs, we perform numerical and experimental tests.  
Fig.~\ref{fig:acc_vs_TOI_s1_0.75_N_200} shows a comparison of simulations and experiments for N=600 performed for scenario s1. The classification accuracy steeply increases for a number of TOIs up to 4, remains stable, and then slightly decreases for larger numbers of TOIs, probably due to the fact that the data base is not large enough to fully train using 10 TOIs. The best experimental accuracy (96.67\%) is obtained using 4 to 6 TOIs. 

Apart from the number of TOIs, the question of what TOIs to concatenate is of particular importance. For example, when concatenating 7 TOIs, the optimal performance is always obtained by including the 10\textsuperscript{th} TOI. The 1\textsuperscript{st} TOI is the second most important TOI, appearing in 95\% of the top 20 results. TOI 9 is in third position (80\%) followed by TOIs 4, 5 and 6 (75\%) and TOI 2 (65\%). 
In other words, TOIs can be ordered depending on their contribution to the classification accuracy. 
Based on these results, a good choice if we use 3 TOI is to  concatenate the last (10\textsuperscript{th} or 9\textsuperscript{th}), the first (1\textsuperscript{st} or 2\textsuperscript{nd}) and middle (4\textsuperscript{th}, 5\textsuperscript{th}, or 6\textsuperscript{th}) TOIs to obtain the best performance.
It's easy to understand why these TOIs are the optimal ones if we think that every TOI represents the states associated to a video frame. Since the reservoir computer has some memory capability, it is best to sample the video sequences in the beginning, in the middle, and in the end. That way, the system is fed with a good representation of the video over its entire length.

\begin{figure}[h]
\includegraphics[width=8cm]{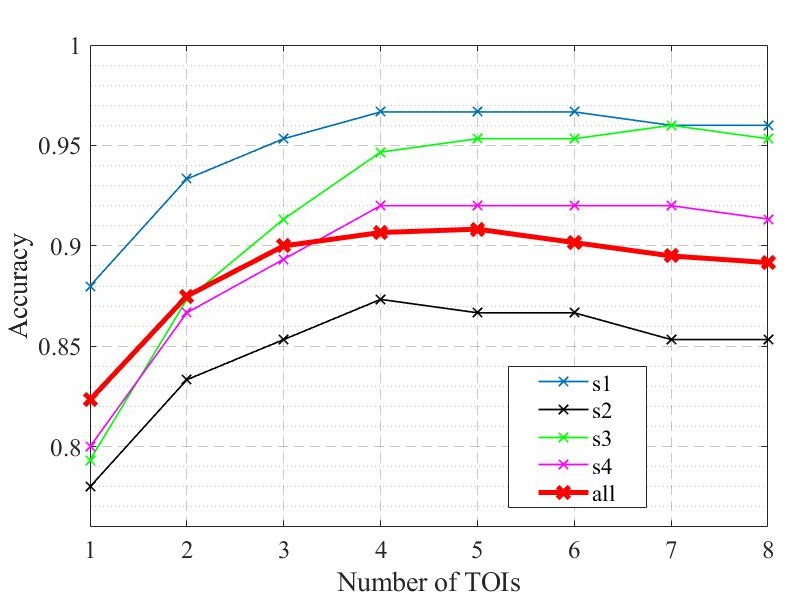}
\centering
\caption{Impact of number of TOIs on performances, for N=600. Experimental results for the complete dataset (red) and for each individual scenario (s1-blue, s2-black, s3-green, s4-magenta). The presented results are for the optimal choice of TOIs.}
\label{fig:acc_vs_TOI_N_600}
\end{figure}

Our experimental results, for each scenario and for the  whole database, for N=600 are shown in Fig.~\ref{fig:acc_vs_TOI_N_600}. Also here it is clear that the optimal number of TOIs lies between 4 and 6. 

\begin{figure*}
\includegraphics[width=16cm]{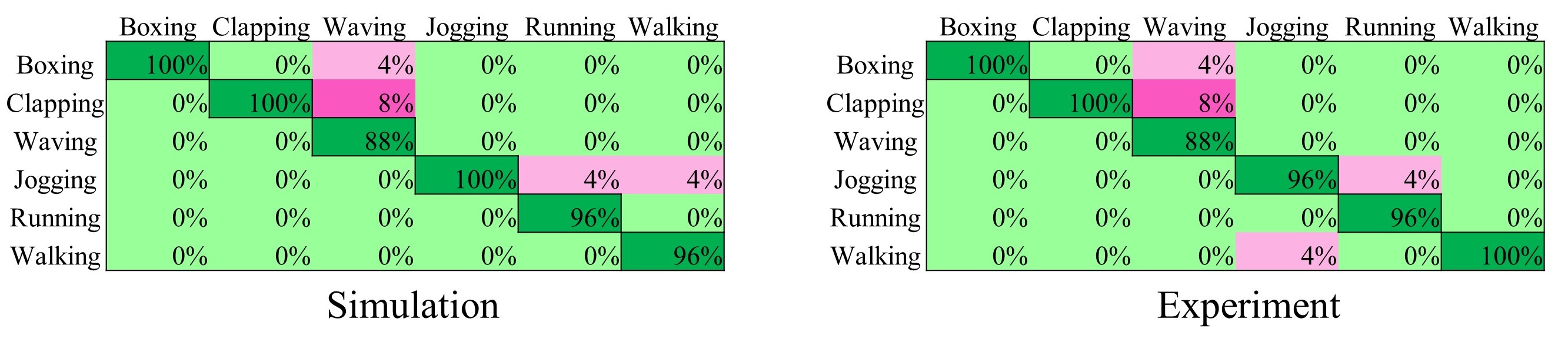}
\centering
\caption{Confusion matrix on test set for N=600, numerical (left) and experimental (right) results for scenario s1.
(Note that the errors in the confusion matrix are multiples of $4\%$ as the test set of scenario s1 contains 150 video sequences, and each class contains $150/6=25$  video sequences; the minimum error is therefore $1/24=4\%$). 
}
\label{fig:confusion}
\end{figure*}

From Fig.~\ref{fig:acc_vs_N} and Fig.~\ref{fig:acc_vs_TOI_s1_0.75_N_200} it is evident that our experimental results are in excellent agreement with the numerical ones. This is not only clear when taking into account the overall accuracy, but also when considering the classification of the individual action class: see the confusion matrices in Fig. \ref{fig:confusion} (for N=600 and scenario 1) which are very similar. 

The enhancement of the performance given by the concatenation of multiple TOIs comes at a price: the training time increases with the number of TOIs. Thus, for a network size of N=600 and the whole dataset, the training time increases from 3.88s (using 1 TOI) to 4.2s (using 7 TOIs) in simulations and from $\sim$ 9 minutes to $\sim$ 12 minutes in experiments. In this case, the amount of additional training time is not too alarming (+8.2\% for simulation and +33\% for experiment) but it can easily explode in case of larger networks where the matrix multiplications become the bottleneck of the computation speed. This aspect has to be taken into account in future when designing new schemes. Additional considerations on the improvement of the experiment speed can be found in Sec.~\ref{sec:conclusion}.

\subsection{Dimensionality Reduction via Principle Component Analysis}\label{sec:PCA}
\begin{figure}[h]
\includegraphics[width=8cm]{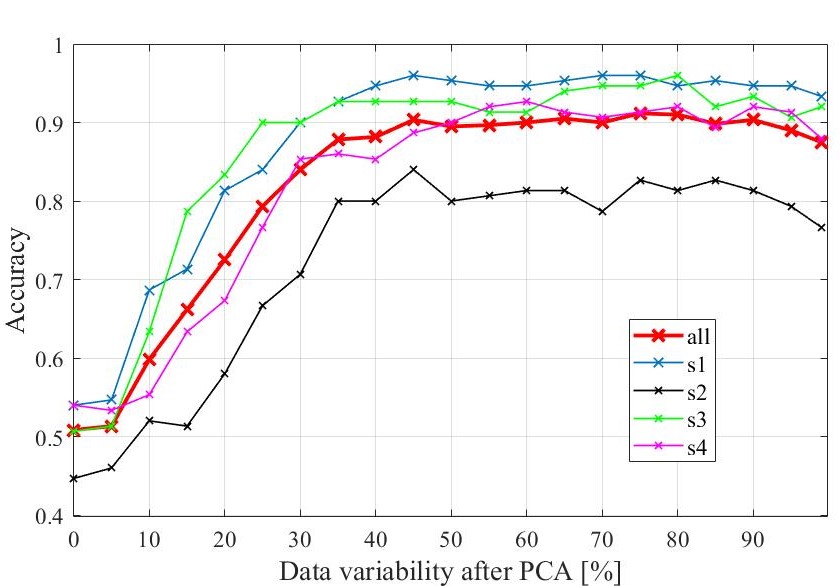}
\centering
\caption{Impact of the data variability kept after PCA, for N=200. Experimental results for the complete dataset (red) and for every individual scenario (s1-blue, s2-black, s3-green, s4-magenta)}
\label{fig:acc_vs_PCA}
\end{figure}

As mentioned in Sec.~\ref{sec:pre-processing}, the last preprocessing step is the input dimensionality reduction using PCA. We use it because too many features can degrade the system performance. A degradation happens when the neural network correlates an output class to some features that do not relate to the class itself but to some noise in the data, due to the finite size of the data base. Further, reducing the input size has also the effect of simplifying computations. For these reasons, we reduce the number of features using PCA by keeping the principal components with the largest eigenvalues, i.e. the ones that account for the most variability in the data. Fig.~\ref{fig:acc_vs_PCA} shows the classification accuracy as a function of PCA data variability. The accuracy increases steadily up to 45\% of the variability and remains stable until around 85\% when it starts to slightly decrease (probably because the database is not large enough to fully train on all the features). Using these results, we set our optimum point at 75\% of variability and use this value through all of this work. The variability of 75\% corresponds to 118 features (out of 1361). 

\subsection{Comparison with Literature }\label{sec:comparison}

In table \ref{table:comparison} we compare our work with some representative results from the vast (hundreds of publications) literature that use the same database. 
We selected works that provide information on the network size and/or the time required for classification, but which employ different feature extraction and classifier methods.

Concerning accuracy, we achieved the highest reported accuracy for prediction of the first (s1) scenario values (96.67\%) (however most publications do not report this) and middle-range accuracy on the whole dataset (90.83\%). Reported results on the whole dataset go from less than 80\% accuracy to essentially perfect (99,3\% accuracy).
Note that the prediction accuracy of our system could be increased further by choosing a larger reservoir size N (Sec.~\ref{sec:reservoirsize}). This could be achieved by increasing the operating frequency or using a longer fiber spool for the optoelectronic reservoir.  

The main focus of this work was to introduce and study a delay based RC as a classifier for HAR. Therefore, we chose a quite simple approach, namely HOG, for feature extraction during the data preprocessing step. HOG was introduced in 2005 \cite{Dalal}. Since then, more advanced feature extraction techniques or combinations  thereof have been published that allow for a considerable increase of the classification accuracy \cite{AbuBakar}. 
The accuracy of our system could thus probably be improved by using these more sophisticated feature extraction methods.

Concerning complexity and processing time, our architecture stands out as being low complexity with only $N=600$ nodes. This low complexity implies fast processing rates. Indeed after training, our architecture processes video frames at rates of 160 frames per second (fps) experimentally and 143 fps numerically. It can therefore classify 25 fps video sequences in real time. 
Note that for the speed analysis we took into account the preprocessing time which for the whole dataset accounts to roughly 29 seconds. But the preprocessing is a small fraction of the time required to run the RC on the whole dataset at the maximum speed of 160 fps (1494 seconds), hence it is almost negligible when calculating the system processing speed. Furthermore, the preprocessing time could be reduced by using a high-end computer. 

Because of its speed, the system could be used to process different video streams at the same time, thus, implementing a multi-channel video processing system. For future works, an additional speed gain of at least one order of magnitude could be reached by attaching an external RAM to the FPGA evaluation board. Indeed opto-electronic reservoir computers based on delay loops can be made considerably faster as demonstrated in \cite{LargerMillion}.
(However if the speed gain was greater than two orders of magnitude the preprocessing time could become the bottleneck of the system). 

Finally, recall that other groups (except for \cite{AntonikHuman} and \cite{Jhuang}) use a classifiers that are not explicitly suited for temporal signals. Therefore, they have to form a descriptor from the extracted features to classify the videos. The search for the optimal descriptor constitutes a whole area of research with several open directions. Reservoir computing as in our case does not require descriptors, thus, simplifying and speeding up the data processing.

\begin{table*}
\resizebox{16cm}{!}{
\begin{tabular}{l|c|c|c|c|c|c}

\multicolumn{1}{c|}
{\textbf{Publication}}  & \textbf{Preprocess} & \textbf{Method} & \textbf{\begin{tabular}[c]{@{}c@{}}Network\\       size\end{tabular}} & \textbf{\begin{tabular}[c]{@{}c@{}}Processing \\      speed\end{tabular}} & \textbf{\begin{tabular}[c]{@{}c@{}}Accuracy \\      (scenario 1)\end{tabular}} & \textbf{\begin{tabular}[c]{@{}c@{}}Accuracy \\      (Full database)\end{tabular}} \\
\hline
Sharif (2017) \cite{Sharif} & LBP + HOG + Haralick + Euclidean distance + PCA & Multi-class SVM & - & 0.5 fps & - & 99.30\% \\
Khan (2020) \cite{Khan} & PDaUM & Deep CNN & \ $\sim$ 195 millions & - & - & 98.30\% \\
Rahman (2014) \cite{Rahman} & Background segmentation + shadow elimination & Nearest Neighbor & - & 12 fps & - & 94.49\% \\
Rathor (2022) \cite{Rathor} & Frame/pose/joints detection + features extraction + PCA & DNN & \ $\sim$ 5000 & 6 fps & - & 93.9\% \\
Shu (2014) \cite{Shu} & Feature extraction with mean firing rate & SNN+SVM & 24000 & - & 95.30\% & 92.30\% \\
Jahagirdar (2018) \cite{Jahagirdar} & HOG+PCA & KNN & - & - & - & 91.83\% \\
Jhuang (2007) \cite{Jhuang} & hierarchy of feature detectors & SVM & - & 0.4 fps & 96\% & 91.60\% \\
Ramya (2020) \cite{Ramya} &  silhouette extraction + distance \& entropy features & NN & 4310 & - & - & 91.4\% \\
\textbf{This work (experimental)} & \textbf{HOG + PCA} & \textbf{photonic RC} & \textbf{600} & \textbf{160 fps} & \textbf{96.67\%} & \textbf{90.83\%} \\
Grushin et al (2013) \cite{Grushin} & space-time interest points + HOF & LSTM & \ $\sim$29000 & 12-15 fps & - & 90.70\% \\
Ji (2010) \cite{Ji} & CNN & 3D CNN & 295458 & - & - & 90.20\% \\
Xie (2014) \cite{Xie} & Star Skeleton Detector & SNN & 10800 & - & - & 87.47\% \\
\textbf{This work (numerical)} & \textbf{HOG + PCA} & \textbf{delay RC} & \textbf{600} & \textbf{143 fps} & \textbf{96.67\%} & \textbf{87.33\%} \\
Liu (2022) \cite{Liu} & spiking coding & LSM (RC + SNN) + EA & 2000 & - & - & 86.30\% \\
Begampure (2021) \cite{Begampure} & resizing + gray scale + normalization & CNN & \ $\sim$ 5 millions & - & - & 86.21\% \\
Antonik (2019) \cite{AntonikHuman} & HOG +PCA & photonic RC & 16384 & 2-7 fps & 91.30\% & - \\
Schuldt (2004) \cite{Schuldt} & local features + HistLF + HistSTG & SVM & - & - & - & 71.83\% \\

\end{tabular}
}
\caption{ Comparison with state-of-the-art literature. SVM: Support Vector Machine. CNN: Convolutional Neural Network. 
SNN: Spiking Neural Network. KNN: K-Nearest Neighbors. LSTM: Long Short-Term Memory. LSM: Liquid State Machine. EA: Evolutionary Algorithm.}
\label{table:comparison}
\end{table*}

\section{Conclusion}\label{sec:conclusion}

In the present work we applied a reservoir computer architecture based on a ring topology with a single nonlinear node to the task of Human Action Recognition. The task we studied, namely Human Actions Recognition using the KTH video-sequence benchmark dataset, is considered to be highly challenging in the field of neuromorphic computing.  

We implemented this architecture both numerically, and using an experimental system consisting in an optolectronic reservoir and a field-programmable gate array. The ring-topology reservoir uses a standard telecom fiber fed with a time-multiplexed and masked input to create the desired number of neurons and influence the reservoir dynamics. The FPGA masks the input signal and interfaces with the reservoir and a PC. The system is reconfigurable and allows to flexibly try out different experimental configurations.     

We introduce a new way to use RC for classification of time series, which we called the Timesteps Of Interest. These allow to increase the size of the output layer without increasing the size of the reservoir itself, and to combine both the short term memory of the reservoir with the longer term memory obtained by keeping relevant TOIs.  We studied the reservoir computers performance as a function of reservoir size, number of TOIs, number of principal components (to reduce the dimensionality of the input during the preprocessing step). 

Contrary to other popular video processing approaches, our system does not require a descriptor formation to transform the temporal dimension into a spatial dimension. This considerably reduces the complexity of the preprocessing stage.

Our experiment reached 90.83\% in prediction accuracy on the whole dataset, and similar results were obtained in numerical simulations. This is in the middle of the range of previous reported results. Our system stands out for its low complexity (only $N=600$ neurons) and speed  (160fps). This exceeds the typical video frame rate of 25 fps by a factor of 6, and implies that this system could process multiple video streams in parallel.   

In conclusion, the results reported here show that reservoir computing, whether implemented numerically or experimentally, is a highly promising approach for video processing, particularly notable for its speed and low complexity. Future works can focus on the processing of video sequences more realistic than the benchmark KTH database. Further improvements in speed and accuracy should be achievable, enabling multi-channel real-time video processing.





\appendix

\section{Subsampling and Keyframes}
\label{app_sub}
Video sequences of the KTH database have different lengths due to different time duration typical of different classes of actions. They span, on average, from 49 frames for the shortest action (running) to 129 frames for the longest (hand waving). This difference in timescales can potentially confuse the classifier or advantage certain action classes to the detriment of the others. 
Further, not all the frames carry equal amounts of information. First, most of the walking, jogging, and running sequences contain several frames at the beginning and at the end where the subject does not appear. Second, for slow-paced motions, consecutive frames in a 25 fps video include little variation in the subject's stance. Therefore, video sequences could in principle be subsampled without noticeable performance degradation. It means only a few frames are required to guess the correct action. It can be seen in Fig.~\ref{fig:KTH} that some action classes, such as boxing or hand waving, can be determined correctly using a only single frame. Other classes, such as running and jogging, are more subtle to distinguish since the single frames are similar between the two classes, and the relevant information is carried by the time relationship between the frames. This idea  has been studied in detail in \cite{Schindler}, where the authors have shown that snippets of 5-7 frames are enough to achieve a performance similar to the one obtainable with the entire video sequence. 
Our approach combines subsampling and selection of keyframes. First, each video sequence is subsampled by a factor of 3 meaning that only one frame out of 3 is kept. Then, from the subsampled sequence, we select 10 frames from its middle part and save them as keyframes for the given sequence. In this way, we achieve that (i) each video sequence has the same number of frames and (ii) we keep the middle of the video sequence where most of the action happens and thus the relevant information is present, while the beginning and the end are discarded without a loss in performance. Sequences shorter than 30 frames – i.e. those containing less than 10 frames after the subsampling – are extended with empty frames (all black pixels) in the beginning.

\section{Silhouette Segmentation and Centering}
\label{app_sil}
Attention mechanisms \cite{Mnih, Ba, Vaswani} have recently shown great success in natural language processing and other fields \cite{Xiao, Ren}. They are effective and efficient because they focus on the principal features of the input data \cite{Lu} instead of blending them without distinction. 
The extraction of binary silhouettes requires the subtraction of the background from 10 keyframes we obtain after subsampling. Several techniques can be found in the literature, such as e.g. Gaussian mixture models \cite{Goyal}, grassfire \cite{Goudelis}, the Self-Organizing Background Subtraction (SOBS) algorithm \cite{Maddalena}, or based on the GLCM features \cite{Vishwakarma}. 
In this work, for the sake of simplicity, we literally removed the background image from video frames, which is the simplest method for background subtraction. This approach has the advantage of being computationally simple and efficient, but can only be applied when videos are recorded with a stable camera so that a reliable background image can be extracted (for instance: surveillance cameras).
This is the situation of most of the KTH dataset videos, except for the ones of scenario 2 which present scale variations. However, since every video was shot against a uniform background, the simple background subtraction produces acceptable results even in this more complicated situation.

\begin{figure}[h]
\includegraphics[width=8cm]{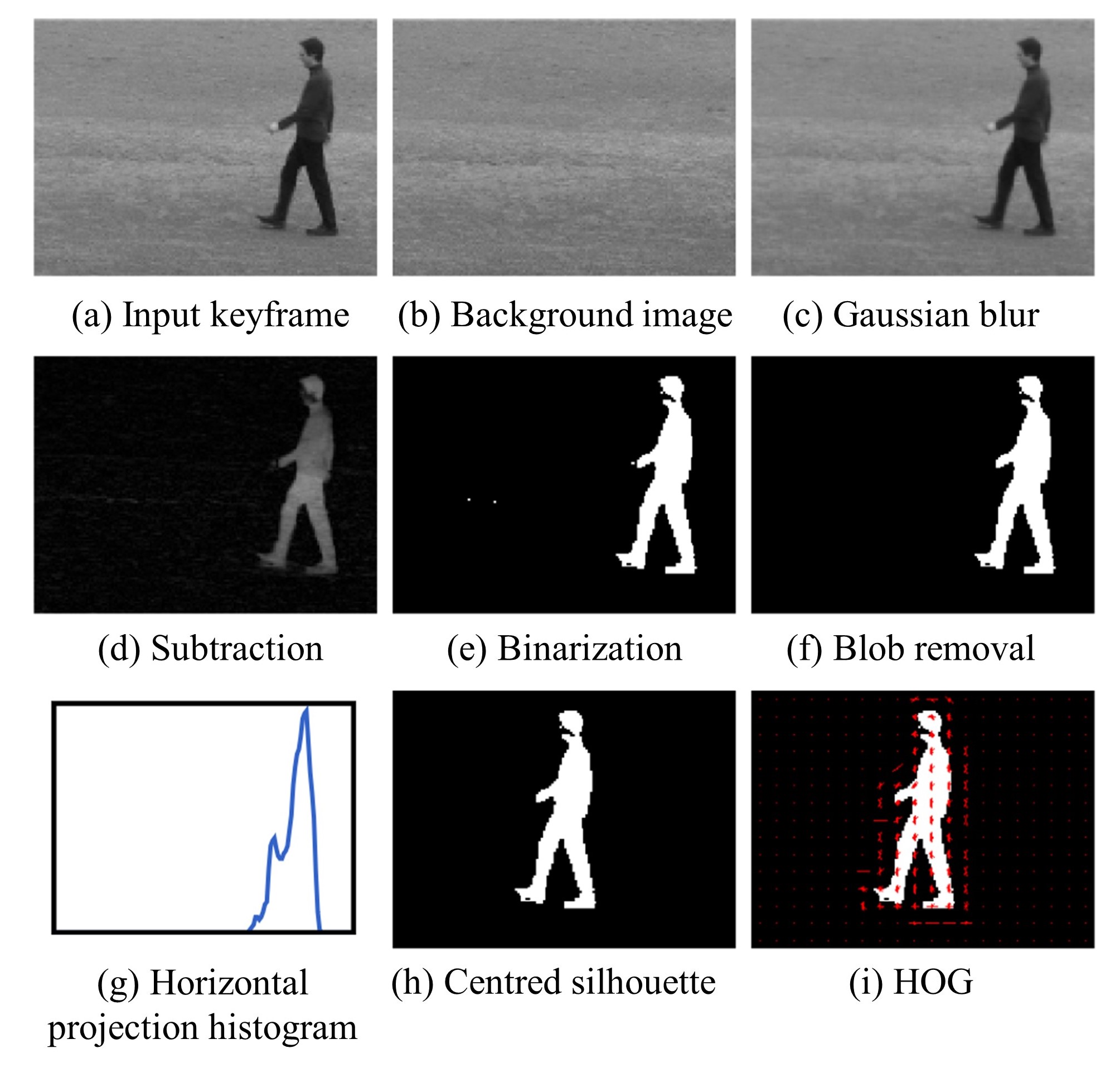}
\centering
\caption{The Preprocessing stage consists in silhouette segmentation (a)-(f), silhouette centering (g)-(h), and feature extraction via HOG(i). After HOG, data are compressed using PCA.}
\label{fig:AppC}
\end{figure}

The silhouette segmentation was implemented in Matlab and consists of the following steps, illustrated in Figs. \ref{fig:AppC}(a)–\ref{fig:AppC}(f):

1) From the raw keyframe (Fig. \ref{fig:AppC}(a)), the background image is selected(Fig. \ref{fig:AppC}(b)). Most of the recordings of the walking, jogging, and running actions start with an empty background since the subject has not yet entered the frame. This frame is the perfect candidate for background subtraction. As for in-place actions (boxing, hand-waving, and hand-clapping), we used a background image from the running sequence of the same subject, in the same scenario.

2) Gaussian smoothing (Fig. \ref{fig:AppC}(c)) is applied to all the keyframes: they are fully blurred using a Gaussian filter,
implemented with the \verb+imgaussfilt+ function in Matlab. Blurring the images allows to reduce the noise and significantly improves the quality of the resulting silhouettes \cite{Stevens}. 

3) Background subtraction (see Fig. \ref{fig:AppC}(d)) is performed pixel-wise.

4) Foreground binarization (Fig. \ref{fig:AppC}(e)). The \verb+imbinarize+ function with a threshold of 0.15 is applied to the resulting images to discriminate the foreground and the background pixels. 

5) Removal of small blobs (Fig. \ref{fig:AppC}(f)). Despite the Gaussian smoothing, background noise such as changes in illumination or shadows sometimes produces small objects, or blobs, in the binarised foreground images (notice the two small white dots in the left-hand side of Fig. \ref{fig:AppC}(e)). To get rid of them, we use the function \verb+bwareopen+ with a threshold of 10, thereby removing any disconnected binary object of 10 pixels or less.

6) The last preprocessing step is the centering stage, illustrated in Fig \ref{fig:AppC}(g) and \ref{fig:AppC}(h). A horizontal projection histogram of the image is traced, to locate the silhouette horizontally; then its maximum is located (we consider the first one in cases of multiple maxima) and shifted to the center of the frame. Finally, the empty space is padded with black columns, to keep the frame size of 160x120 pixels.

\section{Feature Extraction}
\label{app_feature}

In the ML community, numerous feature extraction methods have been used to represent human actions from the input data \cite{AbuBakar}. In general, they belong to one of the two categories: local- or global-part based. The methods based on local features have a long successful history in many applications, given their robustness to disturbances and noise, such as background clutter and illumination changes. Some examples are the scale-invariant feature transform (SIFT) \cite{Lowe}, speeded-up robust features (SURF) \cite{Bay}, spatio-temporal interest points (STIPs) \cite{Li}, dense sampling \cite{Sicre}, and texture based such as local binary patterns \cite{Wang} and GLCM \cite{Haralick}. 
In global-based feature methods, a human shape is derived from a silhouette, a skeleton, or from depth information, and it is then used to capture the action.

In this work, similarly to another previous study \cite{AntonikHuman}, we use Histograms of Oriented Gradients (HOG). 
This method is classified as local features for which we do not take into account the global shapes of the silhouettes extracted from the keyframes. In other words, silhouette extraction is here employed as an attention mechanism. 
The HOG method, introduced by Dalal and Triggs \cite{Dalal}, is based on Scale-Invariant Features transform (SIFT) descriptors \cite{LoweDistinctive}. First, horizontal and vertical gradients are computed by filtering the image with the following kernels \cite{Bahi}: 
\begin{equation}
\begin{aligned}
& G_x = (-1,0,1),  \\
& G_y = (-1,0,1)^T.
\end{aligned}
\label{eq:Gxy}
\end{equation}
Then, magnitude $m(x, y)$ and orientation $\theta(x, y)$ of gradients are computed for every pixel:
\begin{equation}
\begin{aligned}
& m(x,y)=\sqrt{D_x^2+D_y^2}, \\
& \theta(x, y) = \arctan(D_y / D_x).
\end{aligned}
\label{eq:mtheta}
\end{equation}
where $Dx$ and $Dy$ are the approximations of horizontal and vertical gradients, respectively.

Next, the image is divided into small cells (typically 8 × 8 pixels), each of them is assigned a histogram of typically 9 bins, corresponding to angles 0, 20, 40, . . . 160, and containing the sums of magnitudes of the gradients within the cell. This operation provides a compact, yet truthful description of each patch of the image: each cell is described with 9 numbers instead of 64. The procedure is completed with block normalization that allows compensation for the sensitivity of the gradients to overall lighting. The histograms are divided by their euclidean norm computed over bigger-sized blocks. In practice, the computation of HOG features was performed in Matlab, using the built-in \verb+extractHOGFeatures+ function, individually for each keyframe of every sequence, with a cell size of 8 × 8 and a block size of 2 × 2. Given the frame size of 160 × 120 pixels, the function returns 19 × 14 × 4 × 9 = 9576 features per frame. Fig.~\ref{fig:acc_vs_PCA}(i) illustrates the resulting gradients superimposed on top of the silhouette obtained in \ref{app_sil}.
To reduce the input dimensionality, we first remove the zero features, i.e. the gradients equal to zero for all frames in the database, since a large portion of binarized input frames (see Fig. \ref{fig:acc_vs_PCA}(h))) remains empty (black). This simple operation alone reduces the dimensionality approximately by half. 
In the next step, we apply the principal component analysis (PCA) \cite{Pearson}, \cite{Hotelling} based on the covariance method \cite{Smith}. For it, we tune the amount of variability we want to keep. Tab.~\ref{table:features} contains the variability and the number of features generated for the four individual scenarios setting the variability to the maximum (99\%) or the optimal (75\%, see section \ref{sec:PCA}) value. The whole data preprocessing, from the raw video frames to the PCA-compressed inputs, takes 7.25 seconds for every scenario, and so 29 seconds for the whole dataset. Additional considerations on how the preprocessing time affects the overall processing speed can be found in \ref{sec:comparison}.

\begin{table}[h]
\centering
\begin{tabular}{l|c|c}
 Feature set & Variability & Number of features\\
\hline
\hline
Scenario 1 & 99\% & 1361\\
 & 75\% & 118\\
  \hline
Scenario 2 & 99\% & 1288\\
 & 75\% & 109\\
  \hline
Scenario 3 & 99\% & 1437\\
 & 75\% & 134\\
  \hline
Scenario 4 & 99\% & 1475\\
 & 75\% & 131\\
\end{tabular}
\caption{Features statistics generated from different scenarios by applying PCA with maximum variability of 99\% or using optimal variability of 75\%}
\label{table:features}
\end{table}

\section{Reservoir state reset}
\label{app_reset}
A stream of different video sequences in series can lead to the degradation of the reservoir performance as the states related to the initial keyframes can be influenced by the last keyframes of the previous sequence. Therefore, we investigate numerically the reset of the states after each video sequence.  In practice, the reservoir is driven with a null input in between the video sequences, for enough timesteps to ensure that the first nodes related to a sequence are not influenced by the last inputs of the previous sequence. 
The results are shown in Tab.~\ref{table:reset}.
They indicate that the reset considerably reduces the error. We conclude that the state reset can lead to better classification accuracy and use it in our experiment.

\begin{table}[h]
\begin{tabular}{c|c|c|c}
 \multicolumn{2}{c}{Configuration} &  \multicolumn{2}{c}{Performance} \\
 N & TOIs & Reset & No Reset\\
 \hline
 200   & 1 & 77,17\% & 68,83\%\\
 200   & 3 & 80,16\% & 76,33\%\\
 600  & 1 & 82,83\% & 72,00\%\\
 600  & 3 & 85,00\% & 81,33\%\\
\end{tabular}
\caption{Impact of resetting the reservoir states at the beginning of each input video sequence, numerical results.}
\label{table:reset}
\end{table}

\section{Bayesian Optimization}
\label{app_bayes}
Bayesian optimization uses Gaussian Process (GP) regression \cite{Rasmussen} to build a surrogate model of the cost function. This model can be used to search the regions most potential for improvement in the hyperparameter space.
We implemented the Bayesian optimization on the RC model (Eq.~\ref{eq:RC}) using Matlab built-in functions. The GP model generation was done by the function \verb+fitrgp+  using a squared exponential kernel and automatic optimization of the hyperparameters.
We chose \textit{expected improvement} function \cite{AntonikBayesian} as the acquisition function. It evaluates the expected amount of improvement in the objective function ignoring values that cause an increase in the objective.  
All three hyperparameters discussed in Sec.~\ref{sec:hp} were simultaneously optimized within certain intervals. The starting set of observations, i.e. the initial evaluations of the system performance, consisted of 27 samples, comprising all possible combinations of initial values. In our case, these values lie at the extremes and the middle of the hyperparameters intervals. After initial 27 observations, the optimization process was run for 200 observations to obtain the hyperparameters we used for  simulations and experiments in this work. 

\section{Field-Programmable Gate Array (FPGA)}
\label{app_FPGA}

\begin{figure}[h]
\centering
\includegraphics[width=8cm]{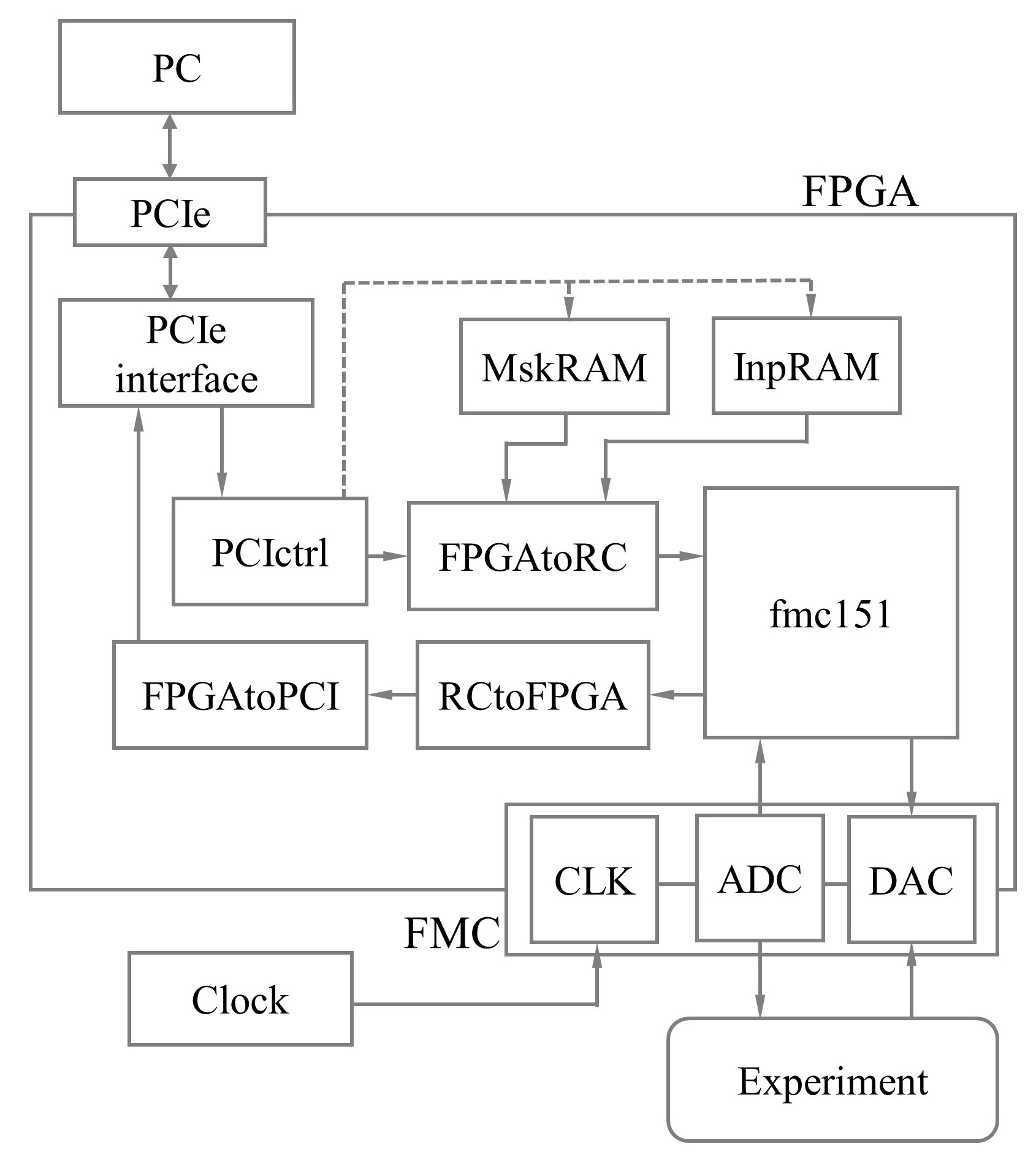}
\caption{FPGA schematic with the entities of the project.}
\label{fig:fpga}
\end{figure}

A Field-Programmable Gate Array (FPGA) is an electronic integrated circuit. Its main feature is reconfigurability which means it is possible to reprogram the FPGA using Hardware Description Languages (HDLs). In our setup, we use Xilinx  Virtex-7 XC7VX485T-FF1761 chip together with the Xilinx VC707 evaluation board. An FPGA Mezzanine Card (FMC) is used to interface with the experiment: we use the 4DSP FMC 151 which contains a dual channel 14-bit ADC and a dual channel 16-bit DAC, with respective bandwidth of 250 MHz and 800 MHz. The board is connected to a PC through a custom PCIe link, which allows communication up to 2 GBps and single data transfer bursts up to 48 GB. In order not to be constrained by the limited frequencies available on-board, the clock tree is driven externally by a Hewlett Packard 8648 A signal generator. For our experiments, we used a frequency of 205 MHz.
The FPGA design is written in standard IEEE 1076-1993 VHDL language \cite{vhdl} and compiled with the Xilinx Vivado 2021 suite. The design is shown in Fig.~\ref{fig:fpga}: each FPGA block represents an entity of the VHDL project. Entities are responsible for various tasks that range from interfacing with the experiment (\textit{FPGAtoRC}, \textit{RCtoFPGA}), implementing and accessing memories (\textit{MskRAM}, \textit{InpRAM}) to communicating with the PC (\textit{FPGAtoPCI}, \textit{PCIctrl}, \textit{PCIe interface}). 
All data and commands for driving the experiment are sent from the PC using a Matlab in-house script.

\section{Role of the reservoir computer in the classification}
\label{app_RC_role}

\begin{figure}[h]
\centering
\includegraphics[width=8cm]{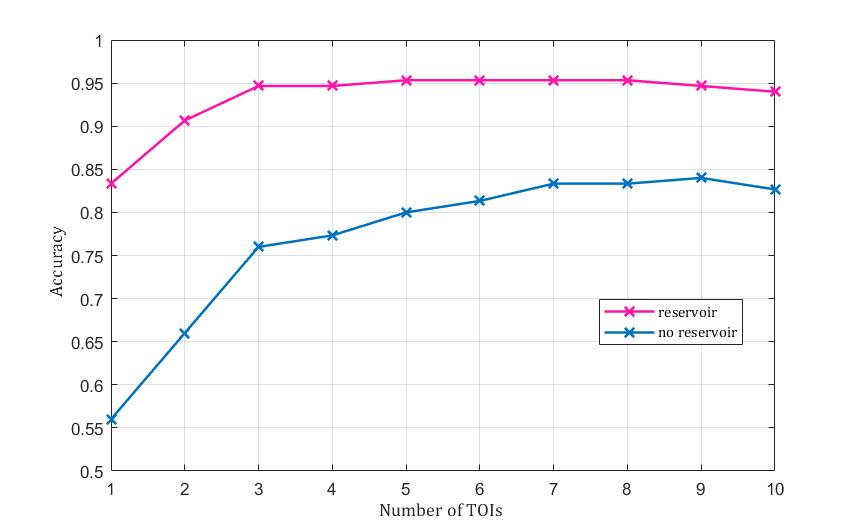}
\caption{Performance of a reservoir with N=200 nodes (pink curve) and performance obtained by taking out the reservoir and performing the linear regression directly on the pre-processed inputs (blue curve).}
\label{fig:RC_role}
\end{figure}

The aim of this section is to quantify the impact of the RC in the classification. As a benchmark for comparison, we evaluate the performance reached if the reservoir is taken out and the linear regression is performed directly on the preprocessed inputs. 
To this end, we performed numerical simulations on scenario 1: the results are shown in Fig. \ref{fig:RC_role}.

We compare the performance of a reservoir with N=200 nodes (pink curve) with the performance obtained by taking out the reservoir and performing the linear regression directly on the pre-processed inputs (blue curve). In both cases, we use the same pre-processed data and the same post-processing for the concatenation of TOIs. 
We observe that for all the possible combinations of TOIs, the configuration with the reservoir outperforms the configuration without the reservoir, with difference in accuracy ranging from 27.3\% (for 1 TOI) to 10.67\% (for 9 TOIs). When considering the best overall performance, i.e.  95,33\% of accuracy with the reservoir and 84\% without the reservoir, we observe a 11.33\% difference in accuracy. In other words, the classification error is reduced from 16\% (without reservoir) to 4.77\% (with reservoir). We can conclude that the reservoir has a major role in the classification.

\section*{Acknowledgements}
The authors would like to express their very great appreciation to Dr. Marina Zajnulina for her valuable and constructive suggestions during the preparation and writing of this research work.

The authors acknowledge financial support from the H2020 Marie Skłodowska-Curie Actions (Project POSTDIGITAL Grant number 860830); and from the Fonds de la Recherche Scientifique - FNRS.
\bibliographystyle{elsarticle-num} 
\bibliography{HAR_article}





\end{document}